\pdfoutput=1

\documentclass[11pt]{article}

\usepackage[]{acl}

\usepackage{times}
\usepackage{latexsym}
\usepackage{bbm}
\usepackage[T1]{fontenc}

\usepackage[utf8]{inputenc}

\usepackage{microtype}

\usepackage{xspace}
\usepackage{algorithmic}
\usepackage{microtype}
\usepackage{booktabs} 
\usepackage{subfigure}
\usepackage{bm}
\usepackage{amsfonts}
\usepackage{soul}
\setul{0.5ex}{0.3ex}
\setulcolor{blue}
\usepackage{makecell}
\usepackage{nicefrac}       
\usepackage{multirow}

\usepackage{microtype} 
\usepackage{epsfig}
\usepackage{diagbox}

\usepackage{framed}
\usepackage{empheq}
\usepackage{array}
\usepackage{enumitem}
\usepackage{mdwlist}
\usepackage{hyperref}

\usepackage{soul}
\usepackage{diagbox}
\usepackage{multirow}
\usepackage{cleveref}

\usepackage[ruled,vlined,commentsnumbered,linesnumbered]{algorithm2e}

\usepackage{bbding}
\usepackage{pifont}

\newcommand{\DataName}{MMDialog}

\title{\DataName: A Large-scale Multi-turn Dialogue Dataset Towards \\ Multi-modal Open-domain Conversation}

\author{Jiazhan Feng$^1$\thanks{\quad Work done during the internship at MSRA.} \quad Qingfeng Sun$^2$ \quad  Can Xu$^2$  \quad Pu Zhao$^2$\\ \quad {\bf Yaming Yang}$^2$ \quad {\bf Chongyang Tao}$^2$ \quad {\bf Dongyan Zhao}$^1$  \quad {\bf Qingwei Lin}$^2$  \\
      $^1$Peking University, Beijing, China \\ $^2$Microsoft Corporation, Beijing, China\\ 
      \texttt{\{fengjiazhan,zhaody\}@pku.edu.cn}\\
      \texttt{\{qins,caxu,puzhao,yayaming,chotao,qlin\}@microsoft.com}}

\begin{document}
\maketitle

\begin{abstract}

Responding with multi-modal content has been recognized as an essential capability for an intelligent conversational agent. In this paper, we introduce the \DataName~dataset to facilitate multi-modal conversation better. \DataName~is composed of a curated set of 1.08 million real-world dialogues with 1.53 million unique images across 4,184 topics. \DataName~has two main and unique advantages. First, it is the largest multi-modal conversation dataset by the number of dialogues by 88x. Second, it contains massive topics to generalize the open domain. To build an engaging dialogue system with this dataset, we propose and normalize two response prediction tasks based on retrieval and generative scenarios. In addition, we build two baselines for the above tasks with state-of-the-art techniques and report their experimental performance. We also propose a novel evaluation metric MM-Relevance to measure the multi-modal responses. Our dataset and scripts are available in \url{https://github.com/victorsungo/MMDialog}.

\end{abstract}

\section{Introduction}


Empowering machines to converse like humans is a long-cherished goal of AI community, and there is growing interest in developing open-domain conversational agents~\cite{li-etal-2017-adversarial,gao2018neural,ghazvininejad2018knowledge,zhou2018emotional}. To usher machines into the world of human knowledge, it is a desirable trait of conversational agents to understand, perceive, and respond appropriately to multi-modality contexts beyond text~\cite{das2017visual,mostafazadeh-etal-2017-image,shuster-etal-2020-image}, which is similar to communicating through messenger tools (e.g., Facebook, WhatsApp, and WeChat) in reality.

\begin{figure}[!htb]
\centering
     \includegraphics[width=0.5\textwidth, scale=1, trim=342 36 291 10,clip]{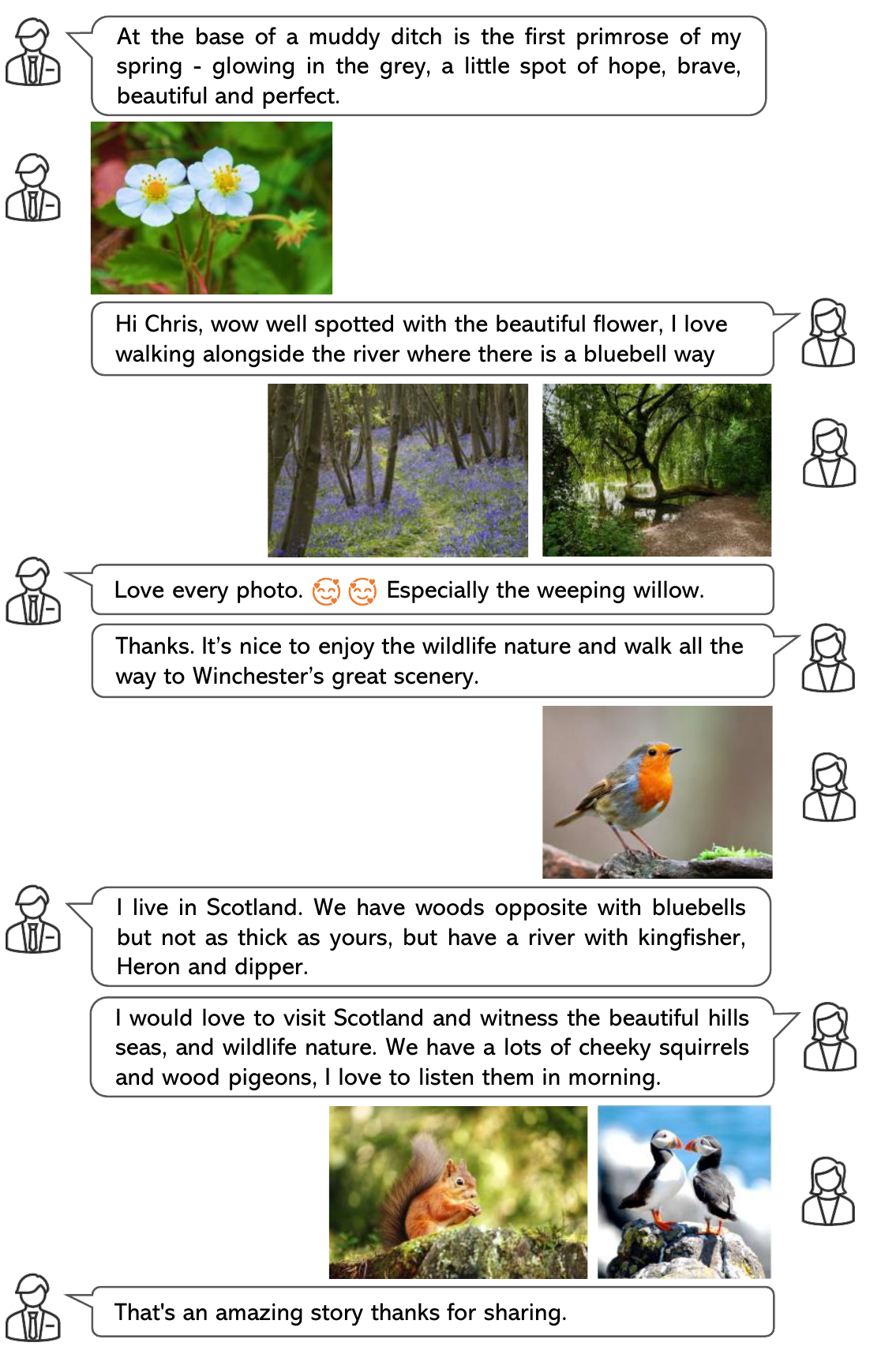}
     \caption{An example of human conversations in our \DataName~dataset. They are talking about scenery and wildlife with both text and various images.}
     \label{fig:intro_case}
\end{figure}

Existing approaches to building multi-modal dialogue systems are primarily data-driven, requiring the collection of a large-scale dataset first. To facilitate this line of research, the community emerges a few dialogue datasets incorporating visual information~\cite{meng2020openvidial,wang2021openvidial,zang-etal-2021-photochat,zheng-etal-2022-mmchat}. For example, Visual Dialog~\cite{das2017visual} is set up for visual question answering involving image inputs. IGC~\cite{mostafazadeh-etal-2017-image} and Image-Chat~\cite{shuster-etal-2020-image} are constructed in a crowd-sourcing method in which annotators are employed to chat about given images. PhotoChat~\cite{zang-etal-2021-photochat} is also built via crowd-sourcing but contains sharing photos in conversations. MMChat~\cite{zheng-etal-2022-mmchat} is collected from real conversations on Chinese social media.

Despite the diversity of multi-modal dialogue corpora, these datasets still have limitations. Firstly, several corpora, including Visual Dialog, IGC and Image-Chat, are derived from crowd-sourcing dialogues talking about given images. The topics of human utterances in a dialogue session are often triggered and grounded by these images, which is inconsistent with our daily communications, where the utterances are not always image-related~\cite{zheng-etal-2022-mmchat}. Secondly, other groups of datasets, such as OpenViDial 1.0/2.0~\cite{meng2020openvidial,wang2021openvidial} and dialogues collected by \citet{lee-etal-2021-constructing}, are not originated from a real multi-modal conversation scenario. The former directly extracts dialogues and their visual contexts from movies and TV series, and the latter replaces some utterances with retrieved relevant images. Both methods artificially construct images from the multi-turn conversation to simulate multi-modal dialogues. Finally, some recently proposed multi-modal dialogue data like PhotoChat and MMChat introduce real human-human conversations. They are still limited by their small scale or lack of domain diversity, impeding the further explorations on multi-modal dialogue modeling.  

To address the aforementioned issues, we present \DataName, a large-scale multi-turn dialogue dataset containing multi-modal open-domain conversations derived from real human-human chat content in social media. \DataName~contains 1.08M dialogue sessions and 1.53M associated images. We elaborately design a series of data filtering processes during the data collection phase. On average, one dialogue session has 2.59 images, which can be located anywhere at any conversation turn.  {Figure~\ref{fig:intro_case} depicts an example of human conversations in our \DataName~dataset.} To the best of our knowledge, this is the first million-scale open-domain multi-modal dialogue corpora. We hope the large amount of dialogues and images can shed light on this line of research.

Furthermore, we define the multi-modal response generation and retrieval tasks based on \DataName~that are essential for building a more engaging multi-modal dialogue agent. We build baseline models and conduct several analyses of their performance. For the generative task, we follow~\citet{sun-etal-2022-multimodal} and implement the models for multi-modal response generation. For the retrieval task, we also propose a CLIP-based dual-encoder for retrieval tasks inspired by~\citet{zang-etal-2021-photochat}. Since in our multi-modal response prediction settings, the modality orders of generated responses may not be aligned with the ground-truth responses. Thus, it is non-trivial to conduct evaluation on cross-modal response elements. To tackle the above challenges, we propose a novel evaluation metric named MM-Relevance, which performs visual-language matching based on the large-scale pre-trained multi-modal CLIP model~\cite{radford2021learning}. Evaluation results on \DataName~demonstrate that our designed baselines can achieve considerable performance on generation and retrieval tasks of both modalities.



To sum up, our contributions are four-fold:

\begin{itemize}
\setlength{\itemsep}{0pt}
    \item We construct a novel multi-turn dialogue dataset \textbf{\DataName}~that contains 1.08M multi-modal open-domain conversations and 1.53M associated images derived from social media and conduct data filtering and post-processing elaborately. To the best of our knowledge, this is the first million-scale multi-turn open-domain multi-modal dialogue corpus.
    \item We propose two benchmark tasks including generative and retrieval scenarios on \DataName~that are essential for building more engaging multi-modal dialogue systems.
    \item We propose a novel evaluation metric \textbf{MM-Relevance} measuring the relevance between generated multi-modal response and ground-truth response. It builds upon the large-scale pre-trained multi-modal CLIP model, which can specifically mitigate the modal misalignment issues.
    \item We design two baselines for corresponding tasks to promote future research on this dataset and achieve considerable performance on generation and retrieval tasks of both modalities. We also give comprehensive analysis to provide more insights into multi-modal dialogue modeling.
\end{itemize}











\section{Related Works}

\subsection{Multi-Modal Dialogue Datasets}
Recently has witnessed a rapid development of visual-language modeling and emerged several multi-modal datasets. The multi-modal datasets include MSCOCO Image Caption dataset~\cite{chen2015microsoft} for image captioning and image generation task; VQAv2 dataset~\cite{goyal2017making} for visual question answering task; SNLI-VE dataset~\cite{xie2019visual} for visual entailment task; RefCOCO~\cite{yu2016modeling}, RefCOCO+~\cite{yu2016modeling} and RefCOCOg~\cite{mao2016generation} for referring expression comprehension task.

Concurrent with the above works, several dialogue-related tasks have also been explored.~\citet{das2017visual} introduced the task of Visual Dialog, which requires an AI agent to hold a meaningful dialogue with humans in natural, conversational language about visual content.~\citet{mostafazadeh-etal-2017-image} proposed IGC, which contains 4K dialogues where each includes an image with a textual description, along with the questions and responses around the image. However, IGC is usually used for evaluation due to its small scale.~\citet{shuster-etal-2020-image} released Image-Chat that is larger than IGC and consists of 202K image-grounded dialogues. However, the above three datasets were created by asking the crowd workers to talk about a shared image to generate the conversation. Therefore, the utterances are often triggered and grounded by these images. In contrast, human daily communication utterances are not always image-related~\cite{zheng-etal-2022-mmchat}, which retain gaps with open-domain multi-modal conversation scenarios. Then, other groups of works proposed to derive the images from the multi-turn conversations:~\citet{meng2020openvidial,wang2021openvidial} constructed OpenViDial 1.0/2.0 by directly extracting dialogues and their visual contexts from movies and TV series.~\citet{lee-etal-2021-constructing} also built a multi-modal dialogue dataset by replacing the selected utterances with retrieved relevant images. However, although these corpora were constructed from open-domain conversations with images, they did not originate from a real multi-modal conversation scenario. Therefore, recently some researchers begin to introduce real human-human conversations.~\citet{zang-etal-2021-photochat} created the first human-human dialogue dataset with photo-sharing acts via crowd-sourcing.~\citet{zheng-etal-2022-mmchat} collected multi-modal dialogues from real conversations on social media. Nevertheless, they were still limited by their small scale or lack of domain diversity, which may hinder further explorations on multi-modal dialogue modeling. To address the aforementioned issue, we make the first attempt to construct a million-scale multi-turn dialogue dataset, namely \textbf{\DataName}, derived from social media and conduct data filtering and post-processing elaborately.

\subsection{Multi-Modal Dialogue Modeling}

Based on the aforementioned multi-modal dialogue datasets, many advanced works have been proposed. Several modeling works~\citep{qi2020two,niu2019recursive,gan2019multi} investigate how to escalate the performance of conversational agents in image-grounded dialogue. Afterward, researchers~\citep{yang2021open,liang2021maria} explore enriching textual expressions of generated dialogue responses through associative vision scenes.~\citet{zang-etal-2021-photochat} proposes two tasks, including photo-sharing intent prediction to predict whether model should intend to share the photo in the next dialogue turn and a dialogue-based image retrieval task to retrieve the most proper photo given the dialogue context. They also propose a dual-encoder model that uses object labels to encode image features, which achieves the best performance among all the models w/o cross-attention mechanisms. However, the authors do not conduct textual response retrieval tasks.~\citet{zheng-etal-2022-mmchat} proposes a multi-modal dialogue generation model based on Seq2Seq architecture, which was proved to be superior to textual Seq2Seq model. However, this model can only generate plain textual responses, which is not in line with the open domain multi-modal response generation scenario. Recently,~\citet{sun-etal-2022-multimodal} make the first attempt to build a multi-modal dialogue response generation model named Divter that can effectively understand multi-modal dialogue context and generate informative text and high-resolution image responses. As advanced works on dialogue systems include retrieval-based methods~\citep{wu2017sequential,zhou2018multi,whang2020effective,li2021small} and generative methods~\citep{li2015diversity,serban2016building,zhang2020dialogpt}. Therefore, we adapt Divter~\cite{sun-etal-2022-multimodal} to our multi-modal response generation settings and extend the dual-encoder~\cite{zang-etal-2021-photochat} to the retrieval-based scenarios as baselines.


\section{Dialogue Creation}
\DataName~is a large-scale multi-turn dialogue dataset towards multi-modal open-domain conversations. It derives from a worldwide social media platform  on which users can converse with each other and share their daily lives messages freely in multiple modalities including plain text, photos, or even videos. We design the data collection process into 3 phases: In the first stage, we extensively manually collect the hashtags commonly used by users and covered as many domains as possible; The second phase starts from the seed hashtags collected before. Specifically we collect all turns with aforementioned hashtags and keep only the turns that contain at least one image, generally we call above turns \textit{anchors} later. Then, for each anchor, we retrieve all the turns that replied to it and the turn it replied to. In the final phase, we also elaborately design a series of data filtering and post-processing steps to eliminate invalid cases and improve the quality of multi-modal dialogues in \DataName. To protect the privacy and security of data, user and platform, \DataName~is released under strict terms for academic people only.

\subsection{Hashtag Collection}
To collect \DataName, we crawl one of the most influential online social platform using its academic available API. To improve the data quality, we consider extracting dialogues with their hashtags (e.g. `\#travel', `\#friends', `\#golf' ), as hashtags tend to show the main topic of the textual utterances and the visual media. Specifically, we manually screen out 4,184 popular hashtags, and each hashtag has at least 1,000 dialogues, in this way our dataset can not only satisfy the properties of open-domain, but also ensure a large scale. We depict the most popular hastags in Figure~\ref{fig:hashtag_case} in Appendix \ref{appendix:dataset}.



\subsection{Multi-modal Conversations Construction}
Then, we leverage the manually collected hashtags as seeds to construct multi-turn dialogues. At first, for each hashtag, we crawl the turns containing corresponding hashtag and only keep those that contain at least one image object (i.e., \textit{anchors}). Obviously, dialogues containing the anchors are the multi-modal multi-turn dialogues we pursue. Then in the same conversation, for each anchor, we look for all the other turns i) that  replied to anchor until reach the leaf node, and ii) that  anchor replied to up to the root node. Moreover, we could recursively follow the chain of replies to recover the entire conversation.



\subsection{Data Filtering and Post-processing}
Since the style of messages posted on social media platforms are widely varied, the initial version of \DataName~contains a lot of invalid, noisy and even harmful conversations, which may hinder the research conducted on this dataset. To tackle the above issue, we design a series of elaborate data filtering processes to filter out those high-quality multi-modal conversations: a) We remove dialogues containing toxic statements with explicit offensive words; b) We ignore and discard dialogues with GIFs and other modalities (such as videos) which cannot be downloaded immediately. We leave this part of research as future work; c) We remove irregular characters from the dialogue content. For example, we do not consider any urls and `@' items (i.e., expression items for mentioning somebody); d) In particular, we convert emojis and hashtags into corresponding natural language forms  to guarantee the coherence of the dialogues;  e) We remove all  self-talking cases (such as replying to themselves for 2 or more consecutive dialogue turns) to enhance the integrity of the conversations;  f) We discard dialogues with incomplete or missing images; g) We only keep the conversations of no less than 3 dialogue turns. We believe that adopting the above data-filtering and post-processing procedure, the final large-scale multi-turn dialogues can be better leveraged  to develop multi-modal open-domain conversation models.





\begin{table}[t!] 
\centering
\resizebox{0.5\textwidth}{!}{
\begin{tabular}{lcc}
\toprule

Statistics  & PhotoChat &  \DataName \\ \midrule
\#Language              & English           & English \\
\#Open-domain           &   \ding{56}       & \ding{52} \\ \midrule
\#Dialogues             & 12.29K            & 1.08M\\

\#Images                & 10.92K            & 1.53M \\
\#Turns                 & 156.10K           & 4.92M \\
\#Topics/Objects        & 89                & 4,184 \\
Avg.  \#Turns per Dialogue  & 12.71         & 4.56  \\
Avg. \#Images per Dialogue & 0.89           & 2.59  \\
Avg.  \#Tokens per Turn & 6.33              & 15.90 \\
\bottomrule

\end{tabular}
}
\caption{Statistics of \DataName~and previous multi-modal dialogue dataset PhotoChat.}
\label{tab:statistics}
\end{table}

\section{Corpus Statistics}

\DataName~consists of 1,079,117 unique dialogues and 1,531,341 images. The statistics of several multi-modal open-domain dialogue corpora are shown in Table~\ref{tab:statistics}. On average one dialogue session has 2.59 images and 4.56 turns, and the images can be located anywhere in any turns of the conversation. We believe that in daily life, people are free to choose any modalities of conversational expressions at any stage of the conversation, and our dialogue data reflect this organizing style. Compared to the recently released multi-modal open domain dialogue dataset PhotoChat~\cite{zang-etal-2021-photochat}, \DataName~enjoys a significantly larger scale of dialogue data and more visual objects, especially that the volume of dialogue sessions has reached million-level. Since conversations originate from a wide range of hashtags  presenting broad domains, the dialogues in \DataName~are open-domain and cover diverse topics, which can shed light on research of multi-modal dialogue modeling. Besides, on average each dialogue turn in \DataName~contains more text tokens than PhotoChat, demonstrating that our proposed data may convey more semantic information in textual utterances.



\begin{figure*}[bht]
\centering
     \includegraphics[scale=2, width=1\textwidth,  trim=50 125 165 200,clip]{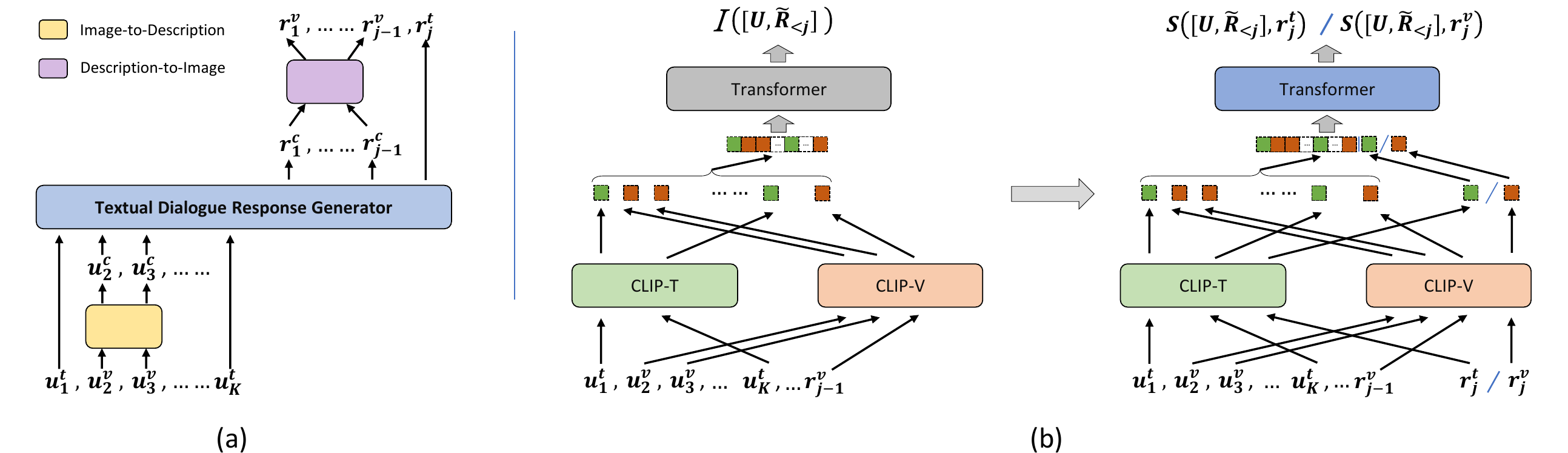}
     \caption{The overview of multi-modal response generation (a) and retrieval (b) baselines.}
     \label{fig:model}
     \vspace{-5mm}
\end{figure*}

\section{Task Definition}
Suppose that we have a multi-modal dialogue dataset $\mathcal{D} =\{(U_i, R_i)\}_{i=1}^n$, where $\forall i \in \{1, ..., n\}$, $U_i$ is the multi-turn dialogue context, $R_i$ is the response regarding to $U_i$. $U_i$ and $R_i$ could contain multi-modal components: textual elements (e.g., utterances) and visual elements (e.g., images). For any $U$ and $R$, we denote $U_i = \{ u^m_k \}_{k=1}^{K}$ and $R_i = \{ r^m_l\}_{l=1}^{L}$ as sequence of multi-modal elements including textual utterances and visual images. $K$ and $L$ are the number of elements in context and response respectively. $m \in \{t, v\}$ indicates the modal type of elements where $t$ represents textual utterances while $v$ signifies visual images. The goal is to learn a multi-modal dialogue model $g$ from $\mathcal{D}$, and thus for any new context $U$, one can predict a multi-modal response $R$ with $g$. 

Since advanced works on pure-text open-domain dialogue systems mainly include retrieval-based  and generative methods. We adapt them to multi-modal scenarios and define the following two tasks that are essential for building a multi-modal open-domain dialogue system:

\paragraph{Task-1: Multi-modal Response Generation}
To generate a multi-modal response $R$, one should learn a multi-modal generation model $P(R|U;\theta)$ with $\theta$ the model parameters. Thus, given a new dialogue context $U$,  following $P(R|U;\theta)$, one can directly synthesize a multi-modal response $\Tilde{R}$ consisting of textual utterances or visual images, or both of them.

\paragraph{Task-2: Multi-modal Response Retrieval}
As for the retrieval-based models, each dialogue example $(U, R)$ additionally provides a series of negative multi-modal elements as distractions. Then we compose the ground-truth textual utterances $\{ r^t_l\}$ in $R$ and the negative examples into an candidate set $C^t = \{ r^{t}_z\}_{z=1}^{Z}$ for text retrieval, where $Z$ is the size of $C$. In the same way, we could also build the image candidates set $C^v = \{ r^{v}_z\}_{z=1}^{Z}$. Thus, the goal of a response retrieval model is to extract an element from a given element candidate set $C^t$ or $C^v$ step by step while predicting each element $r^m_l$. Through such an  retrieval process in an auto-regressive style, we can finally obtain a fully retrieved multi-modal response $\Tilde{R}$.

\paragraph{Response Modal Intent Prediction}
In \DataName, the textual utterances and visual images can be freely located anywhere in the multi-modal response. Therefore, the generation or retrieval order of the modality of response elements is also of great importance for the multi-modal conversation. The intent prediction task aims to predict the order of different modalities in response $\Tilde{R}$ given the dialogue context $U$. Therefore, the intent prediction can be formulated as a classification task: 
\begin{equation}
    \forall j \in [1, J], \mathcal{I}(U, \Tilde{R}_{<j}) \in \{0, 1, 2\}
\end{equation}
where $\mathcal{I}(\cdot,\cdot)$ is the intent prediction model which takes the dialogue context $U$ and previously generated/retrieved response elements $\Tilde{R}_{<j}$ before $j$-th step as inputs and provides the modality of next element. Specifically, the model should predict 0 when $r_j$ is a textual utterance, and 1 when $r_j$ is a visual image. We also define the 2 which indicates that the response $\Tilde{R}$ is completed and the model should stop generating/retrieving new elements.

\section{Evaluation of Multi-Modal Dialogue Tasks}
Since most of the evaluation metrics used for text generation (e.g. BLEU~\cite{papineni-etal-2002-bleu}, ROUGE~\cite{lin-2004-rouge}) or image generation tasks (e.g., FID and IS used in~\citet{ramesh2021zero}) or retrieval (e.g., Recall) can only be evaluated within a single modality. At the same time, the modality orders of elements in a multi-modal dialogue response may not be aligned with the ground-truth response. Thus, it is non-trivial to conduct evaluation on cross-modal response elements. 

In \textbf{Task-1}, we could obtain the BLEU and ROUGE scores by aligning the generated textual parts and those in ground-truth responses from the left. When predicting the next textual element of response, if the model generates no textual element corresponding to the current step of ground-truth element, we could assign the evaluation result of this step to zero value. However, we could not directly adopt the same strategy for metrics such as PPL and FID metrics for textual response generation and image generation tasks respectively as the setting of default zero value is non-trivial. Similarly, we could only compute IS for the generated images. 


In \textbf{Task-2}, we could also compute the Recall scores in similar way. Specifically, we first align the textual (resp., visual) elements in retrieved responses and textual (resp., visual) elements in ground-truth responses from the left. When predicting the next textual (resp., visual) element of response, if the model does not retrieve any textual (resp., visual) element corresponding to the ground-truth response, we could also assign the Recall results of this step to zero values. If the number of previously retrieved textual (resp., visual) elements in $\Tilde{R}_{<j}$ before $j$-th step has reached the number of textual (resp., visual) elements in ground-truth response, the Recall scores at $j$-th step would not be considered, and the model can only retrieve textual (resp., visual) elements from the given negative elements in candidate set $C^t$ (resp., $C^v$). The Recall scores of an example is the average scores of all elements of the same modality. However, the Recall scores can only reveal the model performance of a single modality, and could not comprehensively measure the overall quality of multi-modal responses.

To tackle the evaluation issues in above two tasks, we propose a novel evaluation metric, named MM-Relevance, which performs visual-language matching based on the large-scale pre-trained multi-modal CLIP model~\cite{radford2021learning} for multi-modal dialogue response generation and retrieval tasks. CLIP is trained on a vast corpus of image-caption pairs from Web. It learns to bring the embeddings of both modalities (visual and textual) together via a contrastive objective. Therefore, we utilize this competitive model to assess the relevance between the generated/retrieved responses and the ground-truth responses to mitigate modal misalignment issues. In specific, suppose we obtain a generated or retrieved multi-modal response $\Tilde{R}=\{ \Tilde{r}^m_j\}_{j=1}^{J}$, and the corresponding ground-truth response $R = \{ r^m_l\}_{l=1}^{L}$. We first align the two sequences from the left. Then, the representation vector of each element is obtained by encoding the textual response or visual image through text encoder or image encoder pre-trained by CLIP respectively. We denote the encoded vectors of two responses as: $\Tilde{E}=\{ \Tilde{e}^m_j\}_{j=1}^{J}$ and $E=\{ e^m_l\}_{l=1}^{L}$. Then, we compute the CLIP scores of the two elements position by position until they cannot be aligned:
\begin{equation} 
\begin{aligned}
    \text{MM}_{\text{Rel}}(R, \Tilde{R}) &= \sum_{i=1}^{\min\{L, J \}}(e^m_i)^\text{T} \cdot \Tilde{e}^m_i
\end{aligned}
\end{equation}
In order to penalize the generated/retrieved sequence that is too long or short, we further improve this metric as:
\begin{equation}
\begin{aligned}
    \text{P}_{\text{MM}} &= \frac{\text{MM}_{\text{Rel}}(R, \Tilde{R})}{J} \\ 
    \text{R}_{\text{MM}} &= \frac{\text{MM}_{\text{Rel}}(R, \Tilde{R})}{L} \\
    \text{F1}_{\text{MM}} &= \frac{2\text{P}_{\text{MM}}\text{R}_{\text{MM}}}{\text{P}_{\text{MM}}+\text{R}_{\text{MM}}} \\
\end{aligned} 
\end{equation}
$\text{P}_{\text{MM}}, \text{R}_{\text{MM}}, \text{F1}_{\text{MM}}$ denote soft-precision, soft-recall and soft-F1 score respectively. We take $\text{F1}_{\text{MM}}$ as MM-Relevance. Thus, the relevance degree can now be computed between two modal-misaligned responses $R$ and $\Tilde{R}$. 


With regard to intent prediction, we follow~\citet{zang-etal-2021-photochat} and adopt F1 score as the evaluation metric that measures the accuracy of the model's prediction of the modality order for a dialogue turn. Specifically, we first get the modal sequences of generated/retrieved and ground-truth responses as $\Tilde{M}=\{\Tilde{m}_j\}_{j=1}^J$ and $M=\{m_l\}_{l=1}^L$ respectively. Then, the F1 score can be computed as:
\begin{equation}
    \begin{aligned}
         \text{Match}(M, \Tilde{M}) &= \sum_{i=1}^{\min\{L, J \}}\mathbbm{1} (m_i = \Tilde{m}_i) \\
         \text{P}_{\text{intent}} &= \frac{\text{Match}(M, \Tilde{M})}{J} \\
         \text{R}_{\text{intent}} &= \frac{\text{Match}(M, \Tilde{M})}{L} \\
         \text{F1}_{\text{Intent}} &= \frac{2\text{P}_{\text{intent}}\text{R}_{\text{intent}}}{\text{P}_{\text{intent}} + \text{R}_{\text{intent}}}
    \end{aligned}
\end{equation}
where $\mathbbm{1}$ is an indicator function that has value 1 when $m_i = \Tilde{m}_i$, otherwise 0. In both tasks, $J$ is determined according to the modal sequence of the generative/retrieved response $\Tilde{R}$.

\section{Baselines}
As shown in the Figure \ref{fig:model}, we leverage baseline models to assess \DataName~for the aforementioned two novel multi-modal tasks.

\subsection{Multi-modal Response Generation Model}
We consider to implement the state-of-the-art multi-modal dialogue response generation model Divter (Figure \ref{fig:model}a) proposed by~\citet{sun-etal-2022-multimodal}, which consists of two components: a textual dialogue response generator $\mathcal{G}$ and a description-to-image translator $\mathcal{F}$. 

Specifically, $\mathcal{G}$ takes the dialogue context  $U$ as input, then generates a textual sequence which may contains a textual response  $r^t$ or a textual image description $r^c$ or both of them. Noting that in our settings on \DataName, there may also be several images $u^v$ in multi-turn dialogue context, we thereby replace these images by their descriptions $u^c$ with the help of an image-to-description translation model. In this way, we could concatenate the textual utterances $u^t$ and descriptions into a sequence as the input of $\mathcal{G}$. In addition, we prepend \texttt{[UTT]} and \texttt{[DST]} at the beginning of textual utterance and image description respectively to distinguish them. Then, for a generated   description $r^c$ beginning with \texttt{[DST]}, $\mathcal{F}$ would take them as condition input, and generate a realistic and consistent high resolution image $r^v$ as the real response.

\subsection{Multi-modal Response Retrieval Model}
Inspired by~\citet{parekh-etal-2021-crisscrossed} and \citet{zang-etal-2021-photochat}, we also build a retrieval model $\mathcal{R}$ named DE++ which consists of a modality intent prediction module $\mathcal{R}_\alpha$ and a ranking module $\mathcal{R}_\beta$. As shown in Figure \ref{fig:model}b, before each ranking action, $\mathcal{R}_\alpha$ firstly takes the dialogue context $U$ and previous retrieved response elements $\Tilde{R}_{<j}$ before $j$-th step as inputs and predicts i) the response is completed and model should stop retrieving new elements. or ii) the modality of next elements. If i), the $\mathcal{R}_\alpha$ will take $U$, $\Tilde{R}_{<j}$ as input to predict the intent $\mathcal{I} ([U,\Tilde{R}_{<j}])$; if ii), $\mathcal{R}_\beta$ will calculate the relevance score  $\mathcal{S} ([U,\Tilde{R}_{<j}], r^{m}_j)$. In the same light,  $\mathcal{R}_\beta$ measures all candidates in  $\{r^{m}_z\}_{z=1}^{Z}$ and selects the one with highest relevance score as the final response element at $j$-th step.


Specifically, $\mathcal{R}_\alpha$ and $\mathcal{R}_\beta$ have similar architecture, we adopt CLIP text encoder and CLIP image encoder to represent textual utterance and image respectively. In $\mathcal{R}_\alpha$, we concatenate all the context embeddings with a special learnable \texttt{[CLS]} embedding prepending at the first and feed the embedding sequence into a transformer module to predict the intent. In $\mathcal{R}_\beta$, we prepend the \texttt{[CLS]} embeddings to the concatenated context embeddings sequence or candidate embedding and then feed them into a transformer module separately. After that we can obtain the representation vectors of context and candidate, and compute relevance scores by conducting dot-product of two vectors.





\begin{table}[t!] 
\centering
\resizebox{0.48\textwidth}{!}{
\begin{tabular}{lccc}
\toprule

Statistics  & Training & Validation & Test \\ \midrule
\#Dialogues                 & 1,059,117     & 10,000   &  10,000 \\
\#Images                    & 1,509,284     & 23,812   & 23,772 \\
\#Turns                     & 4,825,053     &  45,382   & 45,801 \\
Avg. \#Turns per Dialogue   & 4.56          & 4.54    & 4.58  \\
Avg. \#Images per Dialogue  & 2.59      & 2.58      & 2.62  \\
Avg. \#Tokens per Turn      & 15.90          & 15.98      & 15.84 \\ \midrule
Avg. \#(Neg. Images) per Dialogue       & - & 999 & 999 \\
Avg. \#(Neg. Utterances) per Dialogue   & - & 999 & 999 \\
\bottomrule

\end{tabular}
}
\caption{Statistics of our training, validation, and test sets.}
\label{tab:statistics_split}
\end{table}



\begin{table*}[ht!] 
\centering
\resizebox{\textwidth}{!}{
\begin{tabular}{lccccccccccc}
\toprule
\multirow{2}{*}{Models} &  \multicolumn{1}{c}{Intent}  & \multicolumn{1}{c}{Image Generation} &  \multicolumn{3}{c}{Textual Response Generation} & \multicolumn{1}{c}{Multi-Modal Generation} \\ 
\cmidrule(lr){2-2}  \cmidrule(lr){3-3} \cmidrule(lr){4-6} \cmidrule(lr){7-7} 
                            &  F1      & IS$\uparrow$  & BLEU-1 & BLEU-2 & ROUGE-L &  MM-Relevance$\uparrow$   \\ \midrule
Divter~\cite{sun-etal-2022-multimodal} &  71.77     &  20.53 $\pm$ 0.50   &   9.44    & 7.45 &   11.19   &  61.85    \\  
\bottomrule
\end{tabular}
}
\caption{Automatic evaluation results of the generative baseline on the test set of \DataName. All numbers except ``IS'' and ``MM-Relevance'' are in percentage.}
\label{tab:main_gen}
\end{table*}



\begin{table*}[ht!]
\centering
\resizebox{\textwidth}{!}{
\begin{tabular}{lcccccccccc}
\toprule
\multirow{2}{*}{Models} &  \multicolumn{1}{c}{Intent} & \multicolumn{3}{c}{Image Retrieval}  & \multicolumn{3}{c}{Textual Response Retrieval} & \multicolumn{1}{c}{Multi-Modal Retrieval}
\\ \cmidrule(lr){2-2} \cmidrule(lr){3-5} \cmidrule(lr){6-8} \cmidrule(lr){9-9}

      &  F1  & R@1   & R@5   & R@10  & R@1   & R@5   & R@10  &  MM-Relevance$\uparrow$ \\ \midrule 
DE++~\cite{zang-etal-2021-photochat}     & 82.69  & 18.23 & 26.99 & 31.73 & 23.07 & 39.21 & 47.05 & 68.91 \\


\bottomrule
\end{tabular}
}
\caption{Automatic evaluation results of the retrieval baselines on the test set of \DataName. All numbers except ``MM-Relevance'' are in percentage.}
\label{tab:main_ret}
\end{table*}





\begin{figure*}[bht!]
\centering  
     \includegraphics[scale=2, width=1\textwidth,  ]{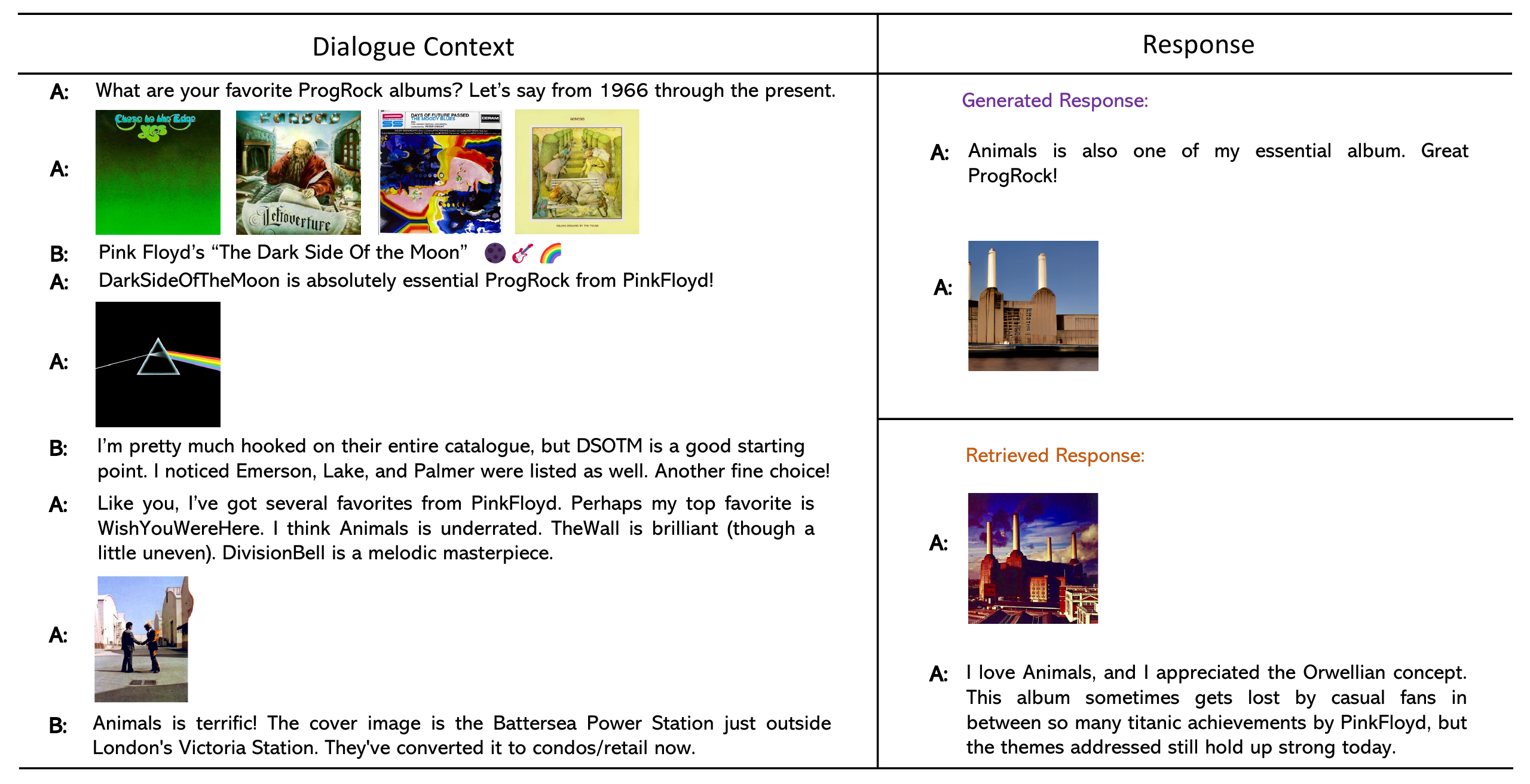}
     \caption{An example of \DataName~test set. \textbf{Left}: the multi-modal dialogue context between ``A'' and ``B''. \textbf{Right}: the multi-modal responses generated or retrieved by our designed baselines.}
     \label{fig:casestudy}
\end{figure*}

\section{Experiments}
Experiments are conducted on \DataName~dataset to assess both our baselines on proposed multi-modal dialogue tasks. We perform response/intent predictions for \textbf{all turns except the first turn} of each dialogue and consider all previous turns as context. 

\subsection{Experimental Setup}
We first sample 10K and 10K dialogue sessions for validation and testing respectively. The detailed statistics are presented in Table~\ref{tab:statistics_split}. For retrieval tasks, we randomly sample 999 negative textual utterances and 999 negative visual images from the same split set for each dialogue, maintaining the total number of candidate elements at 1K. While in training phase, the negative ones are in-batch sampled similar to~\citet{radford2021learning}. For the textual dialogue response generator, we fine-tune DialoGPT~\cite{zhang2020dialogpt} with \textit{transformers} library provided by huggingface\footnote{\url{https://github.com/huggingface/transformers}} using the version ``DialoGPT-medium'' consistent with~\citet{sun-etal-2022-multimodal}. For the description-to-image translator, we implement DALL-E~\cite{ramesh2021zero} using the code of ``mega'' version in \url{https://github.com/borisdayma/dalle-mini}, which also has the same model settings with~\citet{sun-etal-2022-multimodal}. We fine-tune DALL-E mega for one epoch with initial learning rate 1e-8 and mini-batch size of 64. We process all images into 256 $\times$ 256 RGB format for DALL-E. To obtain the description of images in \DataName, we adopt OFA-huge~\cite{wang2022ofa} using the code \url{https://github.com/OFA-Sys/OFA/tree/feature/add_transformers} for image captioning. All version of CLIP models we leveraged in this paper are ``openai/clip-vit-base-patch32'' in \url{https://huggingface.co/openai/clip-vit-base-patch32}. When implementing Divter, we follow the same experimental configuration. As for the retrieval baseline, the representation vectors for both modality are obtained by CLIP model and fixed during training. The transformers used in retrieval tasks consist of 4 Transformer layers with a hidden size of 512 and 8 heads. We train the retrieval models with an initial learning rate of 5e-7 and mini-batch size of 512. For all baselines, early stopping on the validation set is adopted as a regularization strategy and the best model is selected based on the validation performance. The training of both tasks is conducted on 8 Nvidia Tesla A100 80G GPU cards. The BLEU and ROUGE scores are computed by codes in \url{https://github.com/Maluuba/nlg-eval}, while the IS is obtained by \url{https://github.com/toshas/torch-fidelity}.


\subsection{Results of Multi-modal Baselines}
Table~\ref{tab:main_gen} reports the evaluation results of multi-modal response generation baseline. Follow~\citet{sun-etal-2022-multimodal}, we evaluate the textual response generation, image generation and intent prediction tasks. Firstly, we can find that the state-of-the-art model Divter achieves relatively low textual response generation performance (9.44 on BLEU-1 and 11.19 on ROUGE-L) on our proposed \DataName, which validates the difficulty of multi-modal response generation tasks and also demonstrates the necessity of constructing a large-scale multi-modal dialogue dataset for building data-driven models. Secondly, compared with the results on text generation, it is interesting to find that the model achieves better performance on the image generation task and reaches 20.53 on IS. Thirdly, we observe that the baseline achieve a 71.77 F1 score on intent prediction task, indicating that the model has a considerable ability to determine whether to generate text or images during the conversation. Finally, we also leverage the proposed MM-Relevance to evaluate the overall relevance degree between the generated multi-modal dialogue responses and ground-truth ones and our baseline achieves a score of 61.85.

We also conduct the retrieval baselines and show the results in Table~\ref{tab:main_ret}. Our proposed baseline DE++ achieves 18.23\% R@1 and 23.07\% R@1 on image retrieval and textual response retrieval respectively, which demonstrates the capacity of multi-modal retrieval model and the effectiveness of CLIP representation. As for the intent prediction, the F1 score is 82.69 which is superior to the counterpart in generative baseline Divter. Furthermore, we can also find that DE++ obtains a better MM-Relevance score than Divter, which may be attributed to the better intent prediction performance of DE++ and we observe that the alignment of the modality would considerably improve the CLIP matching scores.




\subsection{Case Study}
To further investigate the quality of multi-modal responses predicted by our proposed baselines, we display an example on the \DataName~test data in Figure~\ref{fig:casestudy}. The multi-turn dialogue context between ``A'' and ``B'' is shown in left while the multi-modal responses generated or retrieved by our designed baselines are depicted in right. As we can see, the textual response generated by Divter is coherent with the dialogue context and it can also generate a realistic high-resolution image about the ``Power Station'' in last turn of context, which demonstrates the multi-modal generative capability of our designed generative baseline. As for the retrieval model, our baseline also retrieved a textual response about ``PinkFloyd'' and image on ``Power Station'' semantically related to the dialogue context, which also verifies the effectiveness of retrieval baseline.


\section{Conclusion}
We presented \DataName, a large-scale multi-turn dialogue dataset towards multi-modal open-domain conversation. By extracting turns associated with images and their surrounding contexts from more than 4K topics, \DataName~provides a diverse and open-domain dataset. To facilitate research on building a more engaging multi-modal dialogue system, we define multi-modal response generation and retrieval tasks, and the MM-Relevance metric based on \DataName. We also build baseline models and conduct several analyses of their performance. We believe this can serve as a rich resource to propel research in the multi-modal conversation, for years to help the community propose better methods suited to more scenarios.

\clearpage
\bibliography{custom}  
\bibliographystyle{acl_natbib}

\clearpage

\appendix

\section{Appendix}\label{sec:appendix}

\subsection{Change Log}\label{appendix:changelog}
In order to  protect user privacy and data security, and enable better machine learning technology to model and evaluate \DataName, we made a few adjustments to the dataset and baseline models.
\paragraph{Version-1, \href{https://arxiv.org/abs/2211.05719v1}{11/10/2022}}
\par

\begin{itemize}
\setlength{\itemsep}{0pt}
    \item We release the \DataName.
    \item We propose two benchmark tasks (Multi-modal Response Generation an Multi-modal Response Retrieval) on \DataName.
    \item We propose a novel evaluation metric MM-Relevance measuring the relevance between generated multi-modal response and ground-truth response.
    \item We design two baselines (Divter and DE++) for corresponding tasks.
\end{itemize}

\paragraph{Version-2, \href{https://arxiv.org/abs/2211.05719v2}{11/15/2022}}
\par

\begin{itemize}
\setlength{\itemsep}{0pt}
    \item We update the description of task definition in Section 5 to make it clearer.
    \item We provide more intuitions about the evaluation metrics of multi-modal dialogue tasks in Section 6.
    \item We modify the context and response sequence building mode for Divter model in Section 7. 
    \item We add more detailed settings of experiments and the analyses of evaluation results in Section 8.
    \item We use a hash algorithm to encrypt the key value of the image.
    \item We provide the script for computing the CLIP scores in metric MM-Relevance in GitHub.
    \item We extend the DE++ model from image-only retrieval to text \& image response retrieval.
    \item We re-train the Divter and DE++ models based on the above changes and update the experimental results on \DataName.
\end{itemize}

\paragraph{Version-3, 12/20/2022}
\par

\begin{itemize}
\setlength{\itemsep}{0pt}
    \item We include more evaluation details in Section 6. Specifically, we update the evaluation mode for Task-2 and the corresponding results in Table 4. Now, $J$ is not always equal to $L$ for Task-2, which is consistent with Task-1.

\end{itemize}

\subsection{The Most Popular Hashtags}\label{appendix:dataset}

\begin{figure*}[!tb]
\centering  
     \includegraphics[width=\textwidth]{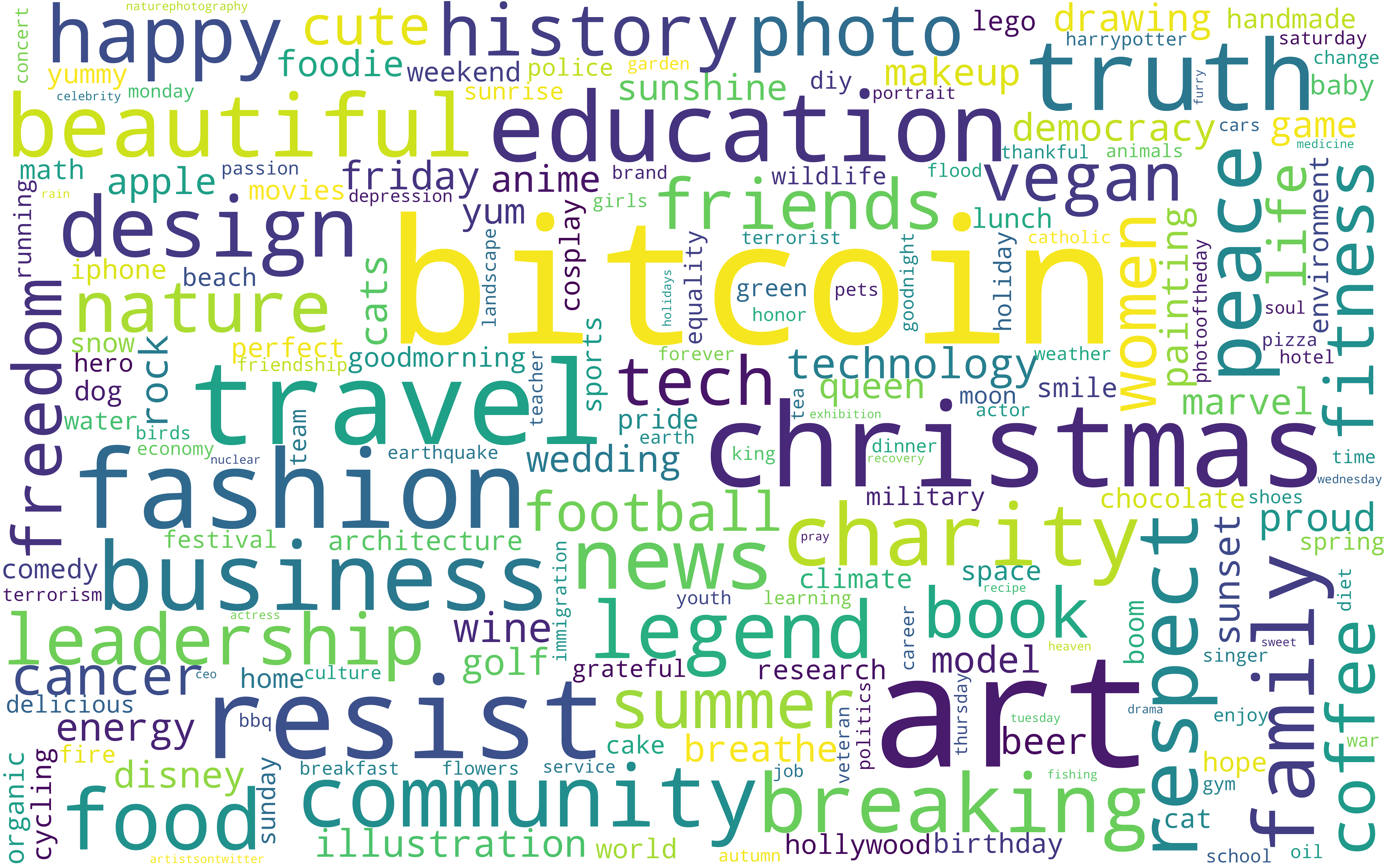}
     \caption{200 most popular hashtags in \DataName~weighted by their frequencies.}
     \label{fig:hashtag_case}
\end{figure*}

We also depict the most popular hashtags in Figure~\ref{fig:hashtag_case}.

\end{document}